\def\year{2020}\relax
\documentclass[letterpaper]{article} 
\usepackage{aaai20}  
\usepackage{times}  
\usepackage{helvet} 
\usepackage{courier}  
\usepackage[hyphens]{url}  
\usepackage{graphicx} 
\usepackage{graphicx}  
\usepackage{amsmath}
\usepackage{amsfonts,amssymb}
\makeatletter
\newif\if@restonecol
\makeatother

\usepackage[linesnumbered,ruled,vlined]{algorithm2e}
\usepackage{algpseudocode}
\usepackage{amsmath}

\nocopyright
\title{Integrated Node Encoder for Labelled Textual Networks }
\author{Ye Ma\textsuperscript{\rm 1}, Lu Zong\textsuperscript{\rm 2} \\[0.8cm] \large \{ye.ma; lu.zong\}@xjtlu.edu.cn }

\begin{document}

\maketitle

\begin{abstract}
Voluminous works have been implemented to exploit content-enhanced network embedding models, with little focus on the labelled information of nodes. Although TriDNR \cite{Pan:2016:TDN:3060832.3060886} leverages node labels by treating them as node attributes, it fails to enrich unlabelled node vectors with the labelled information, which leads to the weaker classification result on the test set in comparison to existing unsupervised textual network embedding models. In this study, we design an integrated node encoder (INE) for textual networks which is jointly trained on the structure-based and label-based objectives. As a result, the node encoder preserves the integrated knowledge of not only the network text and structure, but also the labelled information. Furthermore, INE allows the creation of label-enhanced vectors for unlabelled nodes by entering their node contents. Our node embedding achieves state-of-the-art performances in the classification task on two public citation networks, namely Cora and DBLP, pushing benchmarks up by 10.0\% and 12.1\%, respectively, with the 70\% training ratio. Additionally, a feasible solution that generalizes our model from textual networks to a broader range of networks is proposed.
\end{abstract}

\section{Introduction}
Content-enhanced network embedding, aims at learning continuous vectors for nodes with rich node contents such as texts. These node representations can be directly used in downstream tasks including classification, link prediction, recommendation, etc. To complete or improve these tasks, artificial labels are often added to the node. However, most of the time only a small number of labels could be accessed due to the costly human resources. Therefore, the necessity of making full use of precious human knowledge to automatically annotate unlabelled nodes and enrich their representations emerges.\par
Most textual network embedding models \cite{Yang:2015:NRL:2832415.2832542,Sun2016AGF,tu-etal-2017-cane,Shen_2018,zhang2018diffusion,chen2019improving} focus on learning unsupervised global node embeddings before classifying nodes in a supervised manner, which actually, is deemed as a waste of the valuable labelled information. TriDNR \cite{Pan:2016:TDN:3060832.3060886} learns vectors of node labels which are used to enhance representations of nodes with known labels. However, this approach fails to improve the representations of unlabelled nodes, which explains its weaker performance in the classification task in comparison to other unsupervised node embedding approaches, such as CENE \cite{Sun2016AGF}.
In addition, there are two common disadvantages of network embeddings according to \citeauthor{zhou2018graph} (\citeyear{zhou2018graph}). First, graph embedding often initializes a vector to each node without shared parameters among different nodes in the embedding matrix, which requires a great level of computing resources in the training as the number of nodes grows. Second, it lacks a flexible way to generate representations of new nodes, and often requires the re-construction of a new network to re-train the node vectors. 

To address these problems, this study first uses a shared node/text encoder, whose parameters for different nodes are shared via same words of the input text, to embed all nodes into continuous vectors. Subsequently, we train these node embeddings based on the structural information of the network and node labels. Finally, the trained node encoder, which not only preserves the structural and labelled information, but also extracts the semantic and syntactic features of texts. The node encoder could be used later to generate representations of nodes, including new nodes, by entering their node contents.\par
Our embedding achieves state-of-the-art classification results in different training ratios on two the public textual networks. With the $10\%$ training/labelled ratio, we push the benchmarks up by $2.7\%$ and $4.8\%$ respectively, indicating that even a low percentage of labels improves node representations. Additionally, a feasible solution that generalizes our model from textual networks to a broader range of networks is proposed.

\section{Related Work}
\textbf{Structure-based network embedding:} DeepWalk \cite{Perozzi:2014:DOL:2623330.2623732} employs random walks to sample node sequences whilst skip-gram \cite{mikolov2013efficient} is adopted to learn node representations by regarding nodes as words. LINE \cite{Tang_2015} learns node embedding by taking both the first- and second-order proximity into account. Node2vec \cite{Grover_2016} improves uniform random walks of DeepWalk by adopting biased random walks.\par
\textbf{Textual network embedding:} TADW \cite{Yang:2015:NRL:2832415.2832542} incorporates text information in the network embedding based on the DeepWalk-derived matrix factorization. TriDNR \cite{Pan:2016:TDN:3060832.3060886} trains node representations on three parties, namely node-node, node-word and label-word, so that the trained node vectors simultaneously represent the network-structural, textual and labelled information. CENE \cite{Sun2016AGF} constructs a heterogeneous network by treating documents as nodes and learns node embeddings by second-order proximity. According to CANE \cite{tu-etal-2017-cane}, each node vector is a concatenation of the context-aware text embedding and the structural-based embedding. Context-aware means the text is assigned with different vectors while being connected to different nodes. To improve CANE, WANE \cite{Shen_2018}, DMTE \cite{zhang2018diffusion} and GANE \cite{chen2019improving} are developed to adopt the fine-grained alignment, truncated diffusion maps and global attention via optimal transport, respectively.

\section{Problem Definition}
We introduce basic notations and definitions used in this paper.\par
\textbf{Labelled textual network: }Let $G=(V,E,T,L)$ denote a network, where $V$ is the set of nodes, $E \in (V \times V)$ is the set of edges between nodes, and $T$ and $L$ represent the textual information and the labelled information of nodes, respectively. Each node $v$ denotes a document that is composed of a word sequence $(w_1, w_2, \cdots,w_n)$ with a positional encoding sequence $(p_1,p_2,\cdots,p_n)$, where $w_i,p_i \in \mathbb{R}^d$ and $d$ is the dimension of the node embedding. In addition, $N(v)$ is the network neighborhoods of $v$, and $L(v)$ is the labels of $v$, such as the categorical label. Note that not all nodes have labels.\par
\textbf{Integrated node encoder: }The input of the encoder is the content $T(v)$ of the node $v$ whilst the output is a fixed node vector $\Phi(v) \in \mathbb{R}^d$. The node encoding model simultaneously preserves network structural information and labelled information so that node embedding represents the integrated knowledge.

\section{Algorithm}
\subsection{Overall Framework}
\begin{figure}[h!]
\centering
\includegraphics[scale=0.68]{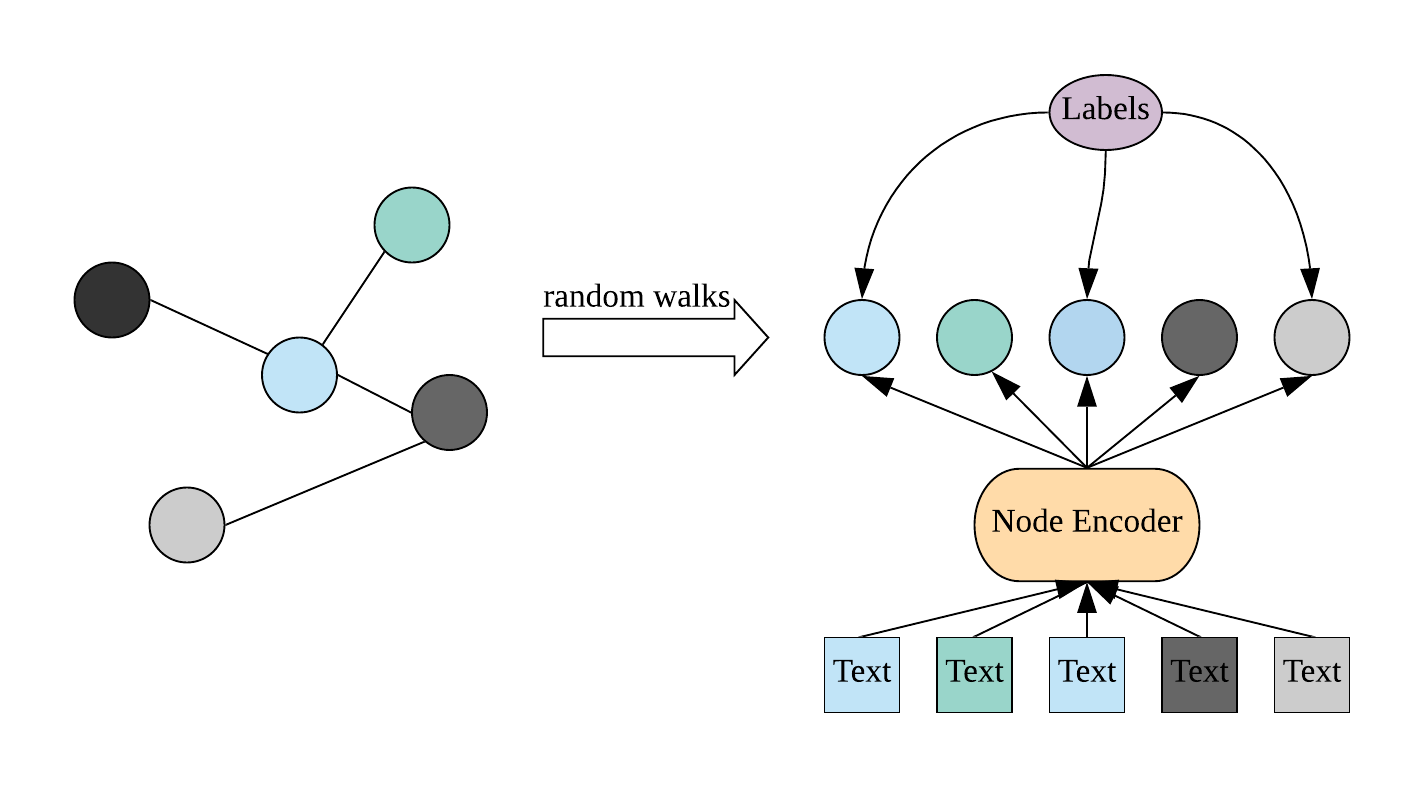}
\caption{Overview of our node encoding model}
\label{fig:view}
\end{figure}

An overview of our node encoding model is presented in Figure \ref{fig:view}. Random walks are employed to sample 
neighbour nodes. The objective of our model is to train an integrated node encoder $\Phi$.
 With such encoder, each node whose content is a document is represented as a continuous vector that preserves both the structural and labelled information of the network. To achieve this goal, the objective is divided into two parts, namely the structure-based objective:
\begin{equation}
    O_s = \sum_{v \in V}logPr(N(v)|\Phi(v)),
\end{equation}
and the label-based objective:
\begin{equation}
    O_l = \sum_{v \in V}\alpha_v \ logPr(L(v)|\Phi(v)),
\end{equation}
\begin{equation}
    \alpha_v=\left\{
             \begin{array}{ll}
             1,& v \ is \ labelled; \\
             0,& v \ is \ unlabelled.
             \end{array}
\right.
\end{equation}
Considering the practical situation that not all nodes have labels, we apply a masking mechanism $\alpha_v$ to our objective function. In practice, we use a special token to denote the label of unlabelled node. No matter what the token is, there is no gradient descent as its gradient is zero after converting the loss to zero. At the same time, this masking mechanism is the key to a fair comparison of our node embedding with other textual network embedding models.

The final objective is assigned to maximize the sum of the above two objectives:
\begin{equation}
    O = O_s + \beta \ O_l
\label{eq:obj}
\end{equation}
where $\beta$ is a constant coefficient. 

In the remaining of this section, a more explicit explanation is given in terms of the architecture of the node encoder, the structure-based \& label-based objectives, as well as their joint training.

\subsection{Node Encoder}

\begin{figure}[h!]
\centering
\includegraphics[scale=0.7]{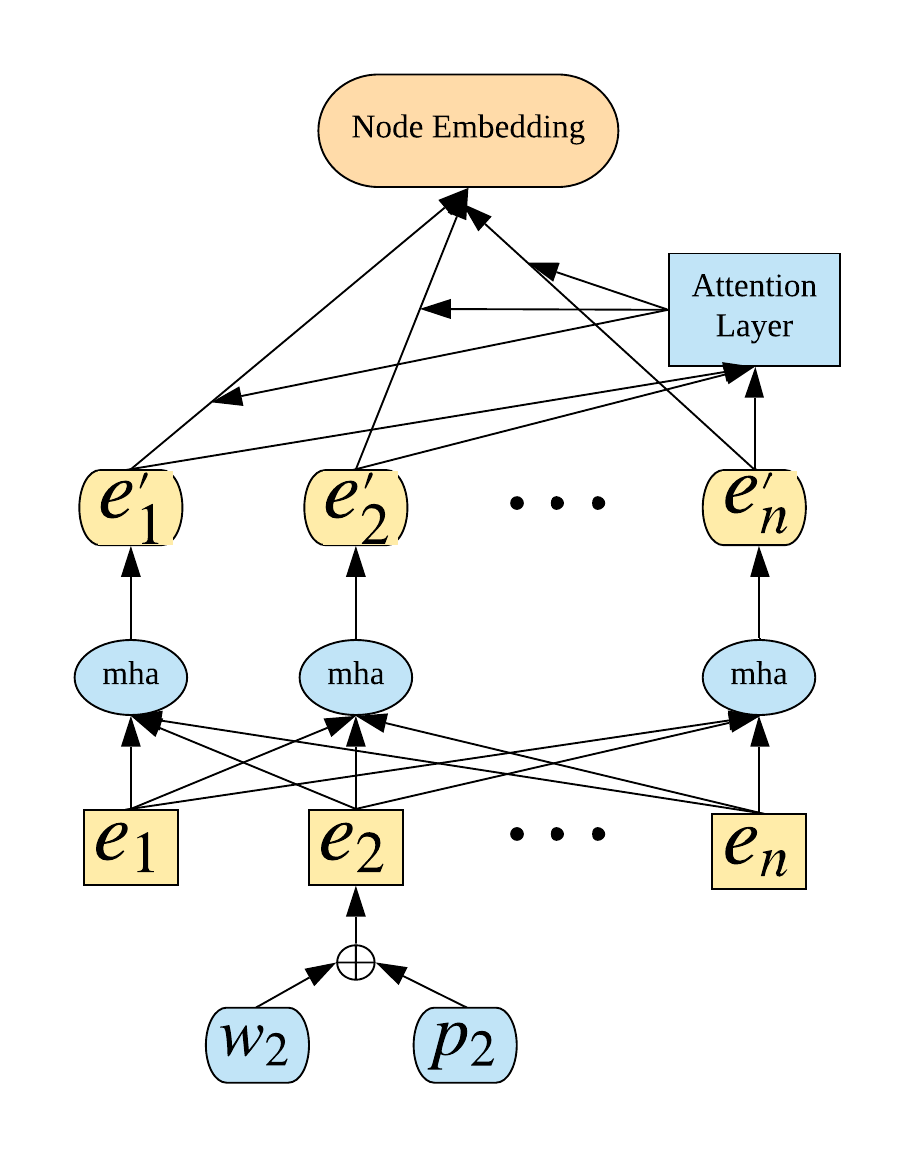}
\caption{Node encoder}
\label{fig:encoder}
\end{figure}

The node encoder converts the document to a fixed vector. As shown in Figure \ref{fig:encoder}, the encoding process is divided into two steps. The first step is to generate the contextual word embedding $e'_i \in \mathbb{R}^d$, whilst the second is to assign weights to each contextual word embedding by an attention layer thus to obtain the node representation, which is the weighted average of the contextual word embeddings.\par
\textbf{Contextual word embedding:} Aims at creating different representations for the same word according to its context. Inspired by the encoding model of Transformer \cite{NIPS2017_7181}, we first add a positional encoding from sine and cosine functions to each word embedding in the right order. To extract the context information of each word, we then compute the intra-attention using the Multi-Head Attention (mha) model which consists of Scaled Dot-Product Attentions. Multi-headers are designed to extract a variety of semantic and syntactic features of a document. For more details on the positional encoding, mha and Scaled Dot-Product Attentions please refer to \citeauthor{NIPS2017_7181} (\citeyear{NIPS2017_7181}).\par
\textbf{Node/text embedding:} The easiest way of computing the text embedding with contextual word embeddings is to use either the add or the average pooling. However, it is practically difficult to mask the padding values in the matrix calculation using these two approaches. Moreover, it is believed that weighted words make more sense than equally important words. Therefore, we adopt another attention mechanism to create node embeddings. The basic architecture of the attention layer is as follows:
\begin{equation}
    h_{v,i} = tanh(W_ee_{v,i}'+b_e) 
\end{equation}
\begin{equation}
    k_{v,i} = \frac{exp(\frac{h_{v,i}^Th_w}{\sqrt{d}})}{\sum_{j=1}^{n}{exp(\frac{h_{v,i}^Th_w}{\sqrt{d}})}} 
\end{equation}
\begin{equation}
    \Phi(v) = \sum_{i=1}^{n}{k_{v,i}e_{v,i}'}
\end{equation}
where $W_e \in \mathbb{R}^{d \times d}, b_e$ and $h_w \in \mathbb{R}^d$ are the trainable parameters. Similar to the Scaled Dot-Product Attention, the same scaling mechanism (i.e. divide each by $\sqrt{d}$) is adopted for more stable gradient. Meanwhile, a mask mechanism is employed on $h_{v,i}^Th_w$ by setting each value in the padding position to $ - \infty$. Additionally, a dropout \cite{JMLR:v15:srivastava14a} is included for $e_i$ and $e'_i$. 
\subsection{Structure-based Objective}
Our structure-based objective is developed in the framework of network embedding based on random walks \cite{Perozzi:2014:DOL:2623330.2623732}. With the assumption of conditional independence, the probability in $O_s$ is factorized as:
\begin{equation}
    logPr(N(v)|\Phi(v))=\sum_{u\in N(v)}logPr(u|\Phi(v)).
\end{equation}
Network neighborhoods $N(v)$ are first sampled by random walks and later sampled again by the context windows after being concatenated into node sequences. To reduce the computational cost, negative sampling \cite{DBLP:journals/corr/MikolovSCCD13} is adopted to approximate the conditional probability.
\begin{equation}
\begin{split}
 logPr(u|\Phi(v)) \approx logPr(1|\Phi(v),u) + \\
\sum_{i=1}^m \mathbb{E}_{u^{'}_i \sim P(v)}[logPr(0|\Phi(v),u^{'}_i)]
\end{split}
\end{equation}
where $u'$ ($u' \neq u$) are sampled $m$ times from the noise distribution $P(v)$. Probabilities of binary classifications are computed by the logistic function.

\subsection{Label-based Objective}
Suppose that each node is assigned with several labels of different types in a given network. In this study, we assume that the node labels are independent of each other, which leads to the conditional probability in $O_l$ written as:
\begin{equation}
    logPr(L(v)|\Phi(v)) = \sum_{l \in L(v)}logPr(l|\Phi(v))
\end{equation}
Since the data sets in the experiment are given with only one type of label, i.e. the node category ($L(v)=c_v$), the conditional probability is simplified as:
\begin{equation}
    Pr(c_v|\Phi(v)) = \frac{exp(\sigma_{c_v}^T \Phi(v))}{\sum_{c\in C}exp(\sigma_c^T \Phi(v))}
\end{equation}
where $\sigma_c$ is trainable and $C$ denotes the set of node classes.

\subsection{Joint Training}

As far as the joint training is concerned, the key problem is that the structure-based objective has more targets than the label-based objective, as each element in the node sequence has several context nodes but only one label and one masking factor. To address this matter, we copy the labels and masking factors so that they have the same number of targets as the structure-based objective does. Algorithm 1 shows the framework of the joint training in our node encoder.\par

\begin{algorithm}
\caption{Joint Training}
\KwIn{$S_n$, the node sequence $[v_1,v_2,\cdots, v_n]$ generated by random walks}
\KwOut{$\Phi$, the trained integrated node encoder}
Initialize $\Phi$ \;
Map node contents into $S_n$, $S_t=[T(v_1),T(v_2),\cdots,T(v_n)]$\;
Map node categories into $S_n$, $S_c=[c_{v_1},c_{v_2},\cdots,c_{v_n}]$\;
Map masking factors into $Sn$, $S_\alpha=[\alpha_{v_1},\alpha_{v_2},\cdots,\alpha_{v_n}]$\;
\For{$t,c,a$ in Zip($S_t,S_c,S_\alpha$)}
{
\For{$u$ in $Context(t)$}
{
$NegativeSampling \ u'$ from $S_t/\{u\}$\;
$AdamOptimizer$ to minimize $-[logPr(u|\Phi(v))+\beta \ a\ logPr(c|\Phi(v))]$ \;
}
}
Return $\Phi$\
\end{algorithm}

From Algorithm 1, we can see the loss function is based on Eq. (\ref{eq:obj}) where $\Phi(v)$ is the output of the node encoder $\Phi$ with the input $t$, and Adam \cite{kingma2014adam} is used to optimize the objective function. Note that in order to give a clearer demonstration of our training framework, we set the batch size to be $1$ in Algorithm 1, whereas the batch size is $64$ by default in the actual operation aiming to speed up the training.

\section{Node Classification}

\subsection{Experimental setup}
All datasets and codes are open source on Github (the link is omitted due to the blind review policy).\par
\textbf{Datasets: }($i$) Cora\url{ https://people.cs.umass.edu/  ̃mccallum/data.ht}, a paper citation network with each node denoting the bibliography data of a paper \cite{McCallum2000}. We use the dataset pruned by \citeauthor{tu-etal-2017-cane} (\citeyear{tu-etal-2017-cane}), which keeps only the abstracts. ($ii$) DBLP\footnote{http://arnetminer.org/citation (V4 version is used)}, a paper citation network similar to Cora. A simplified version built by \citeauthor{Pan:2016:TDN:3060832.3060886} (\citeyear{Pan:2016:TDN:3060832.3060886}) is hired in this study. 
Table \ref{dataset} summarizes the two datasets used in our experiment. Note that out of the $60744$ nodes in total, DBLP has only $17725$ effective nodes that have edges.

\begin{table}[h] 
\centering  
    \begin{tabular}{lcc}
        \hline \hline &Cora & DBLP\\
        \hline
        \#Effective Nodes &2277&17725\\
        \#Edges&5214&52914\\
        \#Node Contents&Abstract&Title\\
        \#Maximum Lengths&410&28\\
        \#Classes&7&4\\
        \hline
        \hline
    \end{tabular}
\caption{Summary of datasets} 
\label{dataset}
\end{table}

\textbf{Parameter setup:} For both Cora and DBLP, we set the coefficient of $O_l$: $\beta = 10$ (After investigating the coefficient, we find that $\beta \ge 3$ outperform smaller $\beta's$ significantly). The dimension of the node embedding is $100$, the dropout rate is $0.5$, the number of attention headers is $2$, the context size is $5$, the walk length is $80$, the batch size is $64$, the epoch is $1$, and negative samples are $5$. Moreover, the number of walks is set to be $3$ for Cora, and $6$ for DBLP.\par

\textbf{The classifier and evaluation criteria: } To train the node encoder, the node set is first shuffled and split into the training and test sets based on the training/labelled ratio. Only labels of the training set are used in the joint training whereas the rest of labels are masked by forcing their losses to zero. The trained node encoder is then used to compute embeddings for the nodes. The support vector machine (SVM) from scikit-learn \cite{scikit-learn} is adopted to make classifications whilst the Macro-f1 score is used to evaluate the classification results. Prior to the SVM, we re-sample the training set to balance the data. The experiment of each train ratio is repeated for at least $5$ times, and the final results are computed as the average Macro-f1 scores.\par
\textbf{Baselines:} We introduce one unsupervised text embedding model Doc2vec \cite{le2014distributed} and nine network embedding models including DeepWalk, LINE, TADW, TriDNR, CENE, CANE, WANE, DMTE, GANE (please refer to Related Work for their briefs) as baselines.\par
The majority of the baselines included use the SVM for classification based on their unsupervised global node embeddings. Due to the supervised nature of SVM, labels are inevitably needed to train the classifier but neglected in the node embedding process, which leads to the loss of the important labelled information. As far as fairness is concerned, our node encoder only makes full use of those labels, which baselines use to train the classifier, at the same time keeps the same proportion of labels as the baselines.
 

\subsection{Results}
The classification results are shown in Table \ref{cora:c} and Table \ref{dblp:c}. Results of the baselines are partially obtained from \citeauthor{chen2019improving} $\dagger$ (\citeyear{chen2019improving}), \citeauthor{Pan:2016:TDN:3060832.3060886} $\ddagger$ (\citeyear{Pan:2016:TDN:3060832.3060886}) and \citeauthor{Sun2016AGF} $\S$ (\citeyear{Sun2016AGF}). We further adjust the hyper-parameters of Doc2vec and DeepWalk for better performance in the classification. In addition, to roundly compare the INE embeddings to the baselines, three versions are included, namely INE-$O_s$ (structural-based: unsupervised version), INE-$O_l$ (label-based), INE-joint (joint training).
\begin{table}[h] 
\centering  
    \begin{tabular}{lcccc}
        \hline \hline Training ratio&10\%&30\%&50\%&70\%\\
        \hline
        LINE$\dagger$ &53.9\%&56.7\%&58.8\%&60.1\%\\
        Doc2vec &71.0\%&74.5\%&76.5\%&76.0\%\\
        DeepWalk&74.9\%&76.6\%&78.7\%&79.4\%\\
        TADW$\dagger$&71.0\%&71.4\%&75.9\%&77.2\%\\
        CANE$\dagger$&81.6\%&82.8\%&85.2\%&86.3\%\\
        DMTE$\dagger$&81.8\%&83.9\%&86.3\%&87.9\%\\
        WANE$\dagger$&81.9\%&83.9\%&86.4\%&88.1\%\\
        GANE$\dagger$&82.3\%&84.2\%&86.7\%&88.5\%\\
        \hline
        INE-$O_s$&76.2\%&82.3\%&83.3\%&84.5\%\\
        INE-$O_l$&49.7\%&80.8\%&94.1\%&97.0\%\\
        INE-joint&\textbf{85.0\%}&\textbf{95.2\%}&\textbf{97.6\%}&\textbf{98.5\%}\\
        \hline
        \hline
    \end{tabular}
\caption{Macro f1 scores on Cora} 
\label{cora:c}
\end{table}

\begin{table}[h] 
\centering  
    \begin{tabular}{lcccc}
        \hline \hline Training ratio&10\%&30\%&50\%&70\%\\
        \hline
        LINE$\ddagger$&42.7\%&43.8\%&43.8\%&43.9\%\\
        Doc2vec &65.3\%&66.7\%&67.4\%&67.4\%\\
        DeepWalk&73.3\%&73.4\%&75.0\%&74.8\%\\
        TADW$\ddagger$&67.6\%&68.9\%&69.2\%&69.5\%\\
        TriDNR$\ddagger$&68.7\%&72.7\%&73.8\%&74.4\%\\
        CENE$\S$&72.8\%&73.7\%&75.0\%&76.3\%\\
        \hline
        INE-$O_s$&74.3\%&76.0\%&76.3\%&76.9\%\\
        INE-$O_l$&21.1\%&21.9\%&42.4\%&41.7\%\\
        INE-joint&\textbf{78.1\%}&\textbf{85.0\%}&\textbf{87.0\%}&\textbf{88.4\%}\\
        \hline
        \hline
    \end{tabular}
\caption{Macro f1 scores on DBLP} 
\label{dblp:c}
\end{table}
 Node vectors generated by the jointly-trained node encoder achieve dominant results in the classification task. Observations are presented as follows:\par
 
\begin{figure*}[h!]
\centering
\includegraphics[scale=0.8]{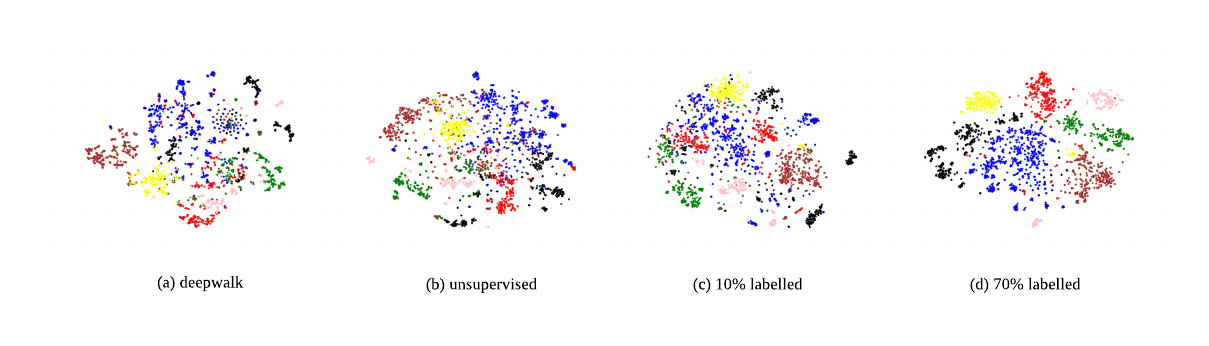}
\caption{Cora visualization}
\label{cora:vis}
\end{figure*}

\begin{figure*}[h!]
\centering
\includegraphics[scale=0.8]{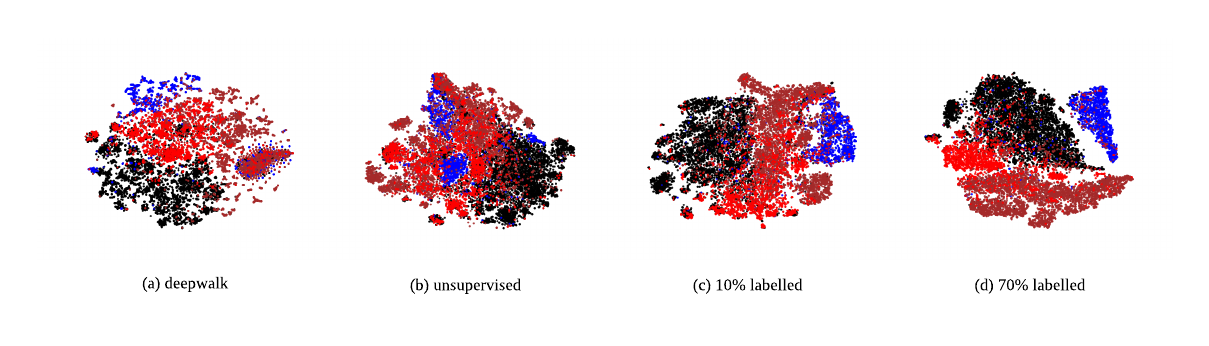}
\caption{DBLP visualization}
\label{dblp:vis}
\end{figure*}

\begin{itemize}
    \item INE-joint significantly outperforms state-of-the-art methods on both Cora and DBLP, indicating labelled information greatly improves the quality of node representations. It should be mentioned that the classification result of CENE on Cora is between CANE and WANE (refer to Figure 3 in \citeauthor{tu-etal-2017-cane} (\citeyear{tu-etal-2017-cane}) and Figure 3(c) in \citeauthor{Shen_2018} (\citeyear{Shen_2018})).
    \item INE-$O_s$ outperforms DeepWalk, showing textual information improves pure structural-based embeddings. At the same time, we find that the improvement on Cora is higher than that on DBLP, due to the lower quality of node contents of the latter. This finding is further supported by the poorer performance of Doc2vec in DBLP than that in Cora, in comparison to DeepWalk.
    \item INE-$O_l$ performs better as the labelled ratio increases on Cora, indicating that the model can produce good classification results by simply relying on the text as long as the training set is adequately large. On the other hand, the poor performance of INE-$O_l$ on DBLP indicates the difficulty in distinguishing the node type based on the node contents of DBLP. In this case, the citation information (i.e. structural information of citation network) plays a major role in the node classification.
\end{itemize} 

\subsection{Investigate labelled ratios}
We investigate the classification results of different labelled ratios on the same training and test set. Labelled nodes are only considered in the training set. As an example, the result of $30\%$ labelled data on a training ratio $10\%$ cannot be computed since the number of labelled data exceeds the number of training data.\par
\begin{table}[h] 
\centering  
    \begin{tabular}{lcccc}
        \hline \hline Training ratio&10\%&30\%&50\%&70\%\\
        \hline
        10\% Labelled &85.0\%&90.7\%&92.8\%&93.3\%\\
        30\% Labelled &-&\textbf{95.2\%}&97.1\%&97.4\%\\
        50\% Labelled &-&-&\textbf{97.6}\%&98.2\%\\
        70\% Labelled &-&-&-&\textbf{98.5}\%\\
       
        \hline
        \hline
    \end{tabular}
\caption{Macro f1 scores of different labelled ratios on Cora} 
\label{cora:d}
\end{table}

\begin{table}[h] 
\centering  
    \begin{tabular}{lcccc}
        \hline \hline training ratio&10\%&30\%&50\%&70\%\\
        \hline
        10\% labelled &78.1\%&78.9\%&78.9\%&79.2\%\\
        30\% labelled &-&\textbf{85.0}\%&84.6\%&84.8\%\\
        50\% labelled &-&-&\textbf{87.0}\%&87.3\%\\
        70\% labelled &-&-&-&\textbf{88.4\%}\\
       
        \hline
        \hline
    \end{tabular}
\caption{Macro f1 scores of different labelled ratios on DBLP} 
\label{dblp:d}
\end{table}

As Table \ref{cora:d} (Cora) and Table \ref{dblp:d} (DBLP) suggest, it is found that given the same training ratio, data with more labels has higher scores. This means that the integrated node encoder preserves more node information and produce higher quality node vectors as the labelled information increases. Moreover, the rate of Macro f1 improvement slows down as the labelled ratio increases. 

\section{Network Visualization}
In order to gain a more intuitive understanding on the advantages of our node representations, we reduce the dimension of the Cora and DBLP embeddings by t-SNE \cite{maaten2008visualizing} (i.e. t-distributed stochastic neighbor embedding). For both datasets, we choose the embeddings produced by DeepWalk, INE-$O_s$ (unsupervised), 10\% labelled INE-joint and 70\% labelled INE-joint for the purpose of comparison. Dimension reduction results are shown in Figure \ref{cora:vis} and Figure \ref{dblp:vis} with the same color indicating the same type of node.\par
 Two cases, namely \textbf{micro clustering} and \textbf{macro clustering}, are defined and discussed for better interpretation of the dimension reduction plots. Specifically, micro clustering means that nodes of the same type (or color) gather in multiple small clusters that may scatter in the graph. A typical example of micro clustering is given by the DeepWalk embeddings. Intuitively, micro clustering indicates that node embeddings captures the network structure information as DeepWalk is solely based on network structure. On the other hand, macro clustering refers to the situation that nodes or small clusters of the same color gather in a certain location close to each other. As the textual and labelled information is included, the phenomenon of macro clustering becomes clearer. In Plots (d) of Figure \ref{cora:vis}, \ref{dblp:vis}, nodes of the same color cluster all together, indicating that our embedding captures the labelled information in a great manner. Nevertheless, Plots (c) and (d) show that macro clustering maintains the characteristics of micro clustering. In particular, the macro clusters consist of micro clusters of the same color, which suggests that our embedding integrates both the structural and labelled information.

\section{Generalized INE}
For textual networks or other networks with node contents, the same framework of the INE could be used by only re-designing the node encoder according to the inputs (i.e. node contents). However, for a broader range of networks without node contents, the input could only be a vector that represents solely the node itself, leading to the two drawbacks according to \citeauthor{zhou2018graph} (\citeyear{zhou2018graph}) (refer to Introduction). To address networks without node contents, we propose to extract node attributes or side information as node contents. The advantages of this technique is twofold. First, the two drawbacks of classic network embedding models \cite{zhou2018graph} are solved. Despite that there could be an infinite number of distinct nodes, a limited number of node attributes that are shared by different nodes could be extracted. Second, this method enriches node vectors with attributes or side information by treating them as encoder inputs. Suppose that a social APP collects its users' data to constructs a social network of users based on their interactions, node contents could be then obtained by analysing and recording their behaviors and characteristics. In this way, node attributes are similar to unordered words that could be inputted into the node encoder. Nonetheless, our model provides a flexible way to incorporate human knowledge (i.e. labels that need to be manually annotated such as customer levels) into the node representations.

\section{Conclusions}
This paper proposes a novel idea to embed networks with node contents. For the textual network, we first design an advanced text encoder to effectively extract semantic and syntactic features. In order to preserve both the structural and labelled information, the node encoder is jointly trained based on structure- and label-based objectives. By making the most of node labels, our experiments show that even a small proportion of node labels improve node representations significantly in the classification task. Further, A shared node encoder, whose inputs are node contents, not only highly saves the computational source in the learning of representations, but also enables the embedding model to generate the embeddings of new nodes. Finally, we discuss the solution to extent our model from textual networks to a wider range of networks. With the generalized framework, network embedding models are given with the capacity to incorporate manually annotated information and node attributes into node representations, at the same time are able to infer representations for new nodes.


\section{Acknowledge}
The authors acknowledge the support of XJTLU Key Programme Special Fund KSF-A-14.

\bibliographystyle{aaai.bst}
\bibliography{aaai}
\end{document}